\documentclass[letterpaper, 10pt, conference]{./support/ieeeconf}
\usepackage{times}
\usepackage[pdftex]{graphicx}

\usepackage{amsmath,amssymb,amsopn,amstext,amsfonts, amsthm}
\usepackage{cancel}
\usepackage[space]{cite}
\usepackage{pdfsync}
\usepackage{balance}
\usepackage{color}
\usepackage{mathtools}
\usepackage[ruled,norelsize, linesnumbered]{algorithm2e}
\usepackage{bm}
\usepackage{diagbox}
\usepackage{float}
\usepackage[outdir=./]{epstopdf}
\usepackage{url}
\usepackage{multirow}
\usepackage{colortbl}
\usepackage{makecell}
\usepackage{adjustbox}
\usepackage{tabularx, booktabs}
\let\labelindent\relax
\usepackage{enumitem}
\usepackage[font=small,skip=2pt]{caption}
\usepackage[linkcolor=black,citecolor=black,urlcolor=black,colorlinks=true]{hyperref}
\usepackage{subcaption}
\usepackage{bookmark}

\theoremstyle{remark}

\SetKw{Continue}{continue}
\makeatletter
\newcommand{\removelatexerror}{\let\@latex@error\@gobble}
\makeatother

\graphicspath{{./figures/}}
\DeclareGraphicsExtensions{.pdf,.png,.jpg,.eps}
\IEEEoverridecommandlockouts
\title{\LARGE \bf Trajectory Prediction with Graph-based Dual-scale Context Fusion}
\author{Lu Zhang$^\dagger$, Peiliang Li$^\ddag$, Jing Chen$^\ddag$ and Shaojie Shen$^\dagger$%
\thanks{$^\dagger$L. Zhang and S. Shen are with the Department of Electronic and Computer Engineering, Hong Kong University of Science and Technology, Hong Kong, China (email: lzhangbz@ust.hk; eeshaojie@ust.hk).
$^\ddag$P. Li and J. Chen are with the DJI Technology Company, Ltd., Shenzhen, China (email: peiliang.li@dji.com; jing.chen@dji.com). This work was supported in part by the HKUST-DJI Joint Innovation Laboratory and in part by the HKUST Postgraduate Studentship.
}%
}

\begin{document}

\maketitle
\thispagestyle{empty}
\pagestyle{empty}

\begin{abstract}
    Motion prediction for traffic participants is essential for a safe and robust automated driving system, especially in cluttered urban environments. However, it is highly challenging due to the complex road topology as well as the uncertain intentions of the other agents. In this paper, we present a graph-based trajectory prediction network named the Dual Scale Predictor (DSP), which encodes both the static and dynamical driving context in a hierarchical manner. Different from methods based on a rasterized map or sparse lane graph, we consider the driving context as a graph with two layers, focusing on both geometrical and topological features. Graph neural networks (GNNs) are applied to extract features with different levels of granularity, and features are subsequently aggregated with attention-based inter-layer networks, realizing better local-global feature fusion. Following the recent goal-driven trajectory prediction pipeline, goal candidates with high likelihood for the target agent are extracted, and predicted trajectories are generated conditioned on these goals. Thanks to the proposed dual-scale context fusion network, our DSP is able to generate accurate and human-like multi-modal trajectories. We evaluate the proposed method on the large-scale Argoverse motion forecasting benchmark, and it achieves promising results, outperforming the recent state-of-the-art methods. We release the code on our project website.\footnote{https://github.com/HKUST-Aerial-Robotics/DSP}
\end{abstract}

\section{Introduction}\label{sec:introduction}

Reasoning about the future motion of other traffic participants in complex traffic is one of the key capabilities to the safe and robust automated driving, especially for the downstream decision-making and motion planning system. However, motion prediction is challenging due to the underlying uncertain and ambiguous intentions of the surrounding agents, especially in complex urban environments. Compared to physics-based and rule-based approaches that depend on a huge amount of domain heuristics, deep learning-based methods show much better performance due to the powerful feature extraction and representation ability for the observed historical trajectory and map information~\cite{lefevre2014survey, mozaffari2020deep}. To obtain the higher efficiency and fidelity, recent state-of-the-art methods~\cite{gao2020vectornet, liang2020learning} leverage vectorized input, wherein agents' trajectories and semantic elements are represented as graphs and then processed by graph neural networks (GNNs). However, these methods usually oversimplify the driving context. For example, trajectories and lanes are simply represented using polylines and further encoded into sparse nodes, making it hard to capture the local information. In the meantime, goal-driven methods~\cite{mangalam2020not, zhao2020tnt, gilles2021home, zeng2021lanercnn} have achieved higher performance on various benchmarks. These methods roughly decompose the trajectory forecasting problem into two sub-tasks, namely, forecasting possible endpoints/goals of target agents, and completing full trajectories conditioned on both context features and predicted goals. Currently, candidate goals are either represented as pixels on rasterized bird-eye-view (BEV) images~\cite{gilles2021home, gilles2021gohome} or generated w.r.t. sparse lane graphs~\cite{zhao2020tnt, zeng2021lanercnn}, making it difficult to consider both high-level and fine-grained driving context. Thus, it remains an interesting and challenging problem on how to design an efficient representation that is capable of capturing multi-scale geometrical and topological features while being compatible with the goal-driven prediction pipeline in a unified way.

\begin{figure}[t]
	\centering
	\includegraphics[width=0.46\textwidth]{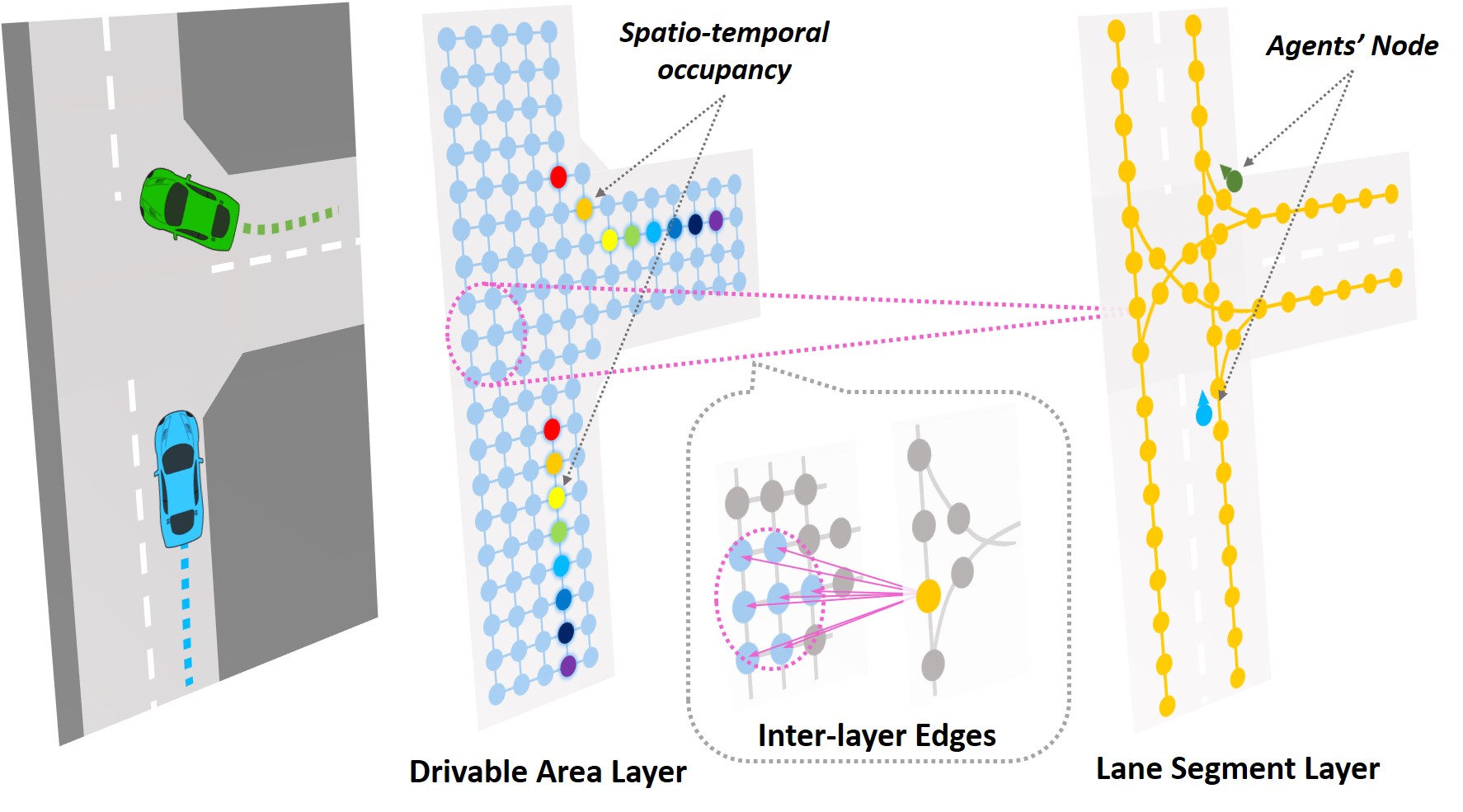}
	\caption{We consider the driving context as a double-layered graph structure: a drivable area (DA) layer and a lane segment (LS) layer, where the DA layer encodes fine-grained local features such as the traversability and spatio-temporal occupancy while the LS layer provides semantic and topological information. Inter-layer connectivities are built according to the Euclidean distance between nodes in the two layers. (Best viewed in color.)}\label{fig:cover}
	\vspace{-1.2cm}
\end{figure}

To address the issue, we present a graph-based trajectory prediction network named the Dual Scale Predictor (DSP). We observe that the driving context in the real world is naturally hierarchical according to the different levels of granularity we are concerned with. As shown in Fig.~\ref{fig:cover}, we formulate the driving context into a double-layered graph structure, which includes a drivable area (DA) layer and a lane segment (LS) layer. The DA is fundamental for the traffic participants, since it physically describes the largest feasible region that the agents can reach. Different from methods using a fixed-size rasterized map to represent the occupancy status, we uniformly draw samples only in the freespace, and connect sampled nodes according to the local traversability. 
Lane structure is another informative clue for motion prediction, especially in urban environments. Thus, in the LS layer, we extract the geometrical and topological information from high-definition (HD) maps, and formulate such information as a sparse but compact graph, wherein nodes are sampled lane segments while edge properties are derived from the lane topology. We leverage GNNs to extract features from the constructed DA and LS graphs, and use attention-based inter-layer graph networks to achieve the dual-scale context fusion. Moreover, the proposed method naturally adapts to the goal-driven prediction framework since the sampled nodes in the DA layer can be directly used as the potential goal candidate. Compared to the existing methods based on rasterized BEV images~\cite{cui2019multimodal, gilles2021home} and sparse lane graphs~\cite{zhao2020tnt, liang2020learning, zeng2021lanercnn}, our DSP is able to consider both lane topology and the fine-grained local features, resulting in better prediction performance. 

Our major contributions are summarized as follows: (1) We present a graph-based hierarchical context representation as well as a graph neural network to achieve efficient feature extraction and aggregation. (2) Following the goal-driven prediction pipeline, we perform endpoint classification over the DA layer, and a multi-modal goal decoder is introduced to further improve the endpoint forecasting. (3) The proposed method achieves state-of-the-art performance on the large-scale Argoverse motion forecasting benchmark.

\section{Related Work}\label{sec:related_work}

Learning-based motion prediction has drawn much attention in recent years. Pioneering works focus on leveraging neural networks to discover the motion pattern under the potential interactions~\cite{alahi2016social, deo2018convolutional, vemula2018social}. In these approaches, the trajectory is usually encoded using recurrent neural networks (RNNs), while convolutional neural networks (CNNs), pooling operators, or attention mechanisms are applied to model interactions among agents. To consider context such as drivable area and HD maps, a common approach is to render the driving environment into multi-channel BEV images and use CNNs to perform feature extraction~\cite{cui2019multimodal, chai2019multipath, zhao2019multi, phan2020covernet}. Cui~\textit{et al.}~\cite{cui2019multimodal} use a target-centric RGB image as input, and semantic elements are rendered using different colors. Since rasterization inevitably causes information loss and ignores the graph topology, context-encoding methods based on the vectorized map data have also been proposed. VectorNet~\cite{gao2020vectornet} takes agents' trajectories and semantic elements as vectorized subgraphs and uses GNNs to encode local features. Then a global interaction graph is applied to perform feature fusion among subgraphs. LaneGCN~\cite{liang2020learning} further utilizes the inherent topology of maps and proposes a novel lane convolution operator, achieving more effective context fusion.

For the multi-modal trajectory prediction task, one training sample has only one ground truth (GT), and regressing all hypotheses leads to the mode collapse issue. To alleviate this problem, a \textit{mode classification} process has been introduced before the trajectory prediction. For example, some works conduct classification on a set of predefined maneuvers~\cite{deo2018convolutional} or possible reference lanes~\cite{zhang2020map, luo2020probabilistic, ding2021epsilon} as an intermediate mode, and then generate final trajectories conditioned on specific modes. Anchor trajectories can also be regarded as a set of driving modes. MultiPath~\cite{chai2019multipath} clusters a set of anchors using offline data, while CoverNet~\cite{phan2020covernet} and PRIME~\cite{song2021learning} generate the candidate trajectory set according to the driving context in an online fashion. Goal-driven methods have also gained popularity in recent years. TNT~\cite{zhao2020tnt} and LaneRCNN~\cite{zeng2021lanercnn} perform endpoint classification and offset regression to generate multi-modal goals, followed by a completion network to get full trajectories. HOME~\cite{gilles2021home} leverages CNNs to encode rasterized BEV images and output a heatmap which represents the probability distribution of the target's goals. Then, final endpoints are sampled over the heatmap. The approach most related to ours is DenseTNT~\cite{gu2021densetnt}, which uses VectorNet to encode sparse context, and utilize an attention module to pass features from sparse nodes to \textit{all} goal candidates, where the candidates are densely sampled w.r.t. a set of promising lanes. An optimization-based post-processing follows to realize accurate goal generation. Compared to the previous methods, our DSP considers both fine-grained and long-range features naturally thanks to our double-layered representation. Moreover, our hierarchical graph structure is well-utilized, which restricts the aggregating operations to the local part of the graph, leading to higher efficiency.

\begin{figure*}[t]
	\centering
	\includegraphics[width=0.94\textwidth]{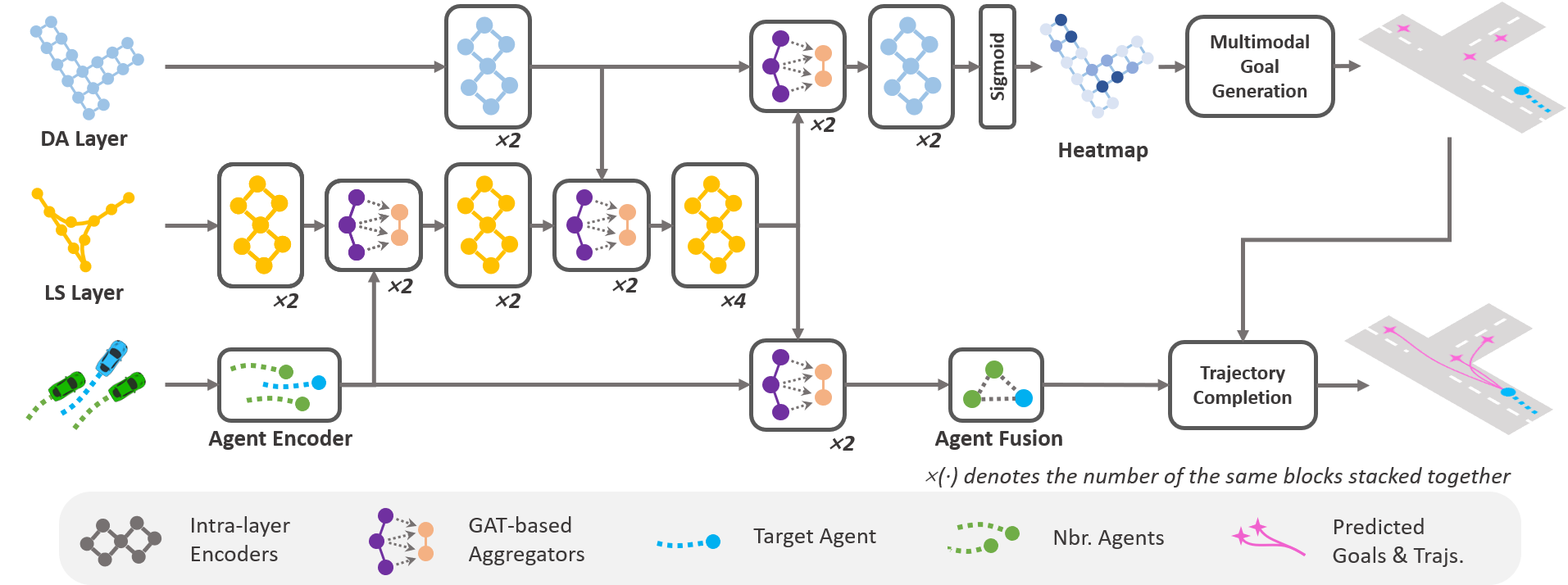}
	\caption{Illustration of the proposed trajectory prediction network. Map features of the DA and LS layers are encoded by their corresponding networks, while motion features are extracted by the agent encoder. GAT-based aggregators are introduced to achieve directed message passing. Intuitively, the DA layer focuses on the fine-grained local context, and the LS layer deals with the long-range feature aggregation, which gives the proposed network a large receptive field as well as the ability to handle detailed structures.}\label{fig:model}
	\vspace{-0.5cm}
\end{figure*}

\section{Methodology}\label{sec:method}
\subsection{Problem Formulation}
The task of a prediction module is to accurately predict the future states of the target agent $\mathbf{y}=[y^1,...,y^H]$ up to a fixed prediction horizon $H$, given the observed history trajectories of the target agent $\mathbf{x}=[x^{-T+1},...,x^{0}]$ and its neighbors $\mathbf{z}=[z^{-T+1},...,z^{0}]$ as well as the map information $\mathcal{M}$, where $T$ is the frames of the observed history. We follow the basic idea of goal-driven methods~\cite{zhao2020tnt, zeng2021lanercnn, gilles2021home} that decompose the prediction problem into endpoint prediction and trajectory completion. Therefore, the original prediction formulation can be written as
\begin{equation*}
    p(\mathbf{y}|\mathbf{c}) \approx \sum_{g_i\in\mathcal{G}}{p(\mathbf{y}|g_i,\mathbf{c})p(g_i|\mathbf{c})},
\end{equation*}
where $\mathbf{c}=(\mathbf{x},\mathbf{z},\mathcal{M})$ is the driving context, $\mathcal{G}$ is a finite set of candidate goals, $p(g_i|\mathbf{c})$ is for endpoint forecasting and $p(\mathbf{y}|g_i,\mathbf{c})$ is for trajectory completion conditioned on predicted goals and context features. Since most of the uncertainty comes from the target's intention, which is essentially unobservable, once the target's endpoint is determined, the trajectory can be easily generated with high accuracy~\cite{zhao2020tnt}. In this paper, we mainly focus on how to obtain representative multi-modal goals, and we show a simple network is capable of achieving satisfactory trajectory completion results.

\subsection{Graph-based Context Representation}
As we mentioned in the previous sections, current goal-driven methods are either based on heatmap prediction~\cite{mangalam2020not, gilles2021home} or classification and offset regression on sparse anchors~\cite{zhao2020tnt, zeng2021lanercnn}. It is difficult for them to consider both fine-grained endpoint information as well as long-range features in a unified framework. For example, rasterized images naturally lose the lane topology, and information loss is inevitable due to the rendering process. Sparse anchors such as the center-point of a small lane segment may contain an ambiguous meaning since different driving behaviors may end in the same segment, especially for low-speed scenarios. 

To address this problem, we propose a graph-based double-layered driving context representation, which includes a DA layer that focuses on local traversability and an LS layer that describes the long-range dependencies. Specifically, for the DA layer, we uniformly sample (resolution: 1.0 $m$) in the drivable area with the target agent as the center. Nodes are connected with their Moore neighborhood if no static obstacle lies between them. Apart from the 2-D positional information, we incorporate the spatio-temporal occupancy status into the node features as well. Additional $T$ channels are added for each node, and a Boolean value is set as \texttt{True} once any of the agents is within a predefined range of the node at the corresponding timestamp. The spatio-temporal occupancy provides more detailed motion features compared to raw trajectories. For the LS layer, we uniformly resample the raw centerlines from HD maps, obtaining a set of short lane segments. We utilize the sampled lane segments (resolution: 2.0 $m$) as the nodes of the LS graph with the 8-dimensional node feature containing the 2-D position, tangent direction and semantic flags (e.g., turning direction and flags for traffic control). Following the formulation in \cite{liang2020learning}, nodes are connected with different types of edges according to the topology of the HD maps, including predecessor, successor, left neighbor and right neighbor. Moreover, inter-layer connectivity is also indispensable for dual-scale feature fusion. Note that the DA and LS layers are from the same map data and always spatially overlapped, therefore, it is reasonable to build the inter-layer edges according to the Euclidean distance between nodes from different layers. An illustration of the proposed representation is shown in Fig.~\ref{fig:cover}.

\subsection{Network Overview}
Based on the proposed hierarchical context representation, we present a network as shown in Fig.~\ref{fig:model}. Firstly, motion features are extracted by a 1-D convolution-based encoder, while map data for the DA and LS layers are encoded using input networks, which lift the feature dimensions to $d_{\text{da}}=32$ and $d_{\text{ls}}=128$, respectively. Then, we leverage two kinds of GNNs as their encoders since the graph structures of the DA and LS layers are different. We design a feature aggregator based on graph attention networks (GATs)~\cite{velivckovic2017graph}, which is able to perform message passing on directed graphs effectively. Motion features and map features from the DA layer are passed successively to the LS graph. After encoding the long-range dependencies on the LS layer, features are sent back to the DA layer and agent features. We next perform the endpoint classification on the DA layer and utilize a multi-modal goal decoder to get the final goals. In the end, the predicted trajectories are generated by a simple trajectory completion module conditioned on the goal hypothesis and the fused agent feature. The detailed design is elaborated in the following sections.

\subsection{Agent Feature Encoding}
Observed trajectories of the target agent and its neighbors are all considered as the input of the agent encoder. To better describe the motion information, we use a 5-dimensional vector for each state on a trajectory, including the position, the tangent direction, and a padding flag if the observation on this timestamp is missing. Then, the trajectory features are passed through a network based on 1-D convolutional layers similar to the implementation in~\cite{liang2020learning}. Note that, in this step, trajectory features are processed independently without considering the interaction between agents. We denote the agent encoder as a mapping $f_{\text{agt}}:\mathbb{R}^{N_{\text{agt}}\times T\times 5}\rightarrow \mathbb{R}^{N_{\text{agt}}\times d_{\text{agt}}}$, where $N_{\text{agt}}$ is the number of agents in the scenario and $d_{\text{agt}}=128$ is the channel size of the generated agent feature.

\subsection{Graph Feature Aggregation}\label{sec:intra}
According to the previous design, the graph structures of the DA and LS layers are different. For example, nodes of the DA graph are connected according to the local traversability, while nodes of the LS layer are linked based on the topological relationship. Additionally, the edges of the DA graph are identical, while different types of edges are involved in the LS graph to represent various connectivities. Focusing on these different aspects, we leverage two kinds of GNNs for encoding the DA and LS graphs. 

\begin{figure}[t]
	\centering
    \begin{subfigure}{0.48\textwidth}
         \centering
         \includegraphics[width=\textwidth]{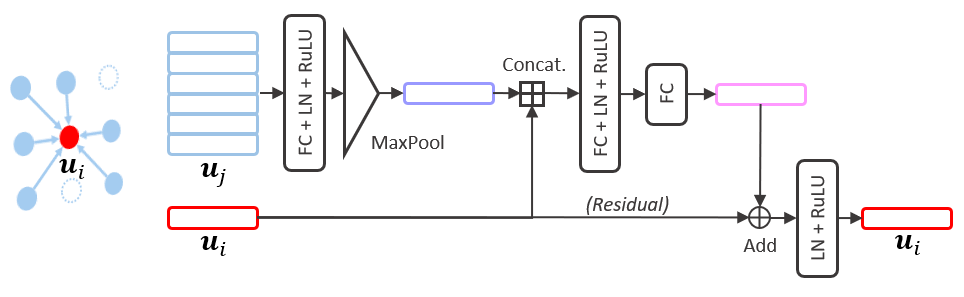}
         \caption{Our purpose is to pass messages from neighboring nodes (blue) to the target node (red). Dashed circles denote no available node at that position. Permutation-invariant aggregation is achieved via a max-pooling operation.}
         \label{fig:da_aggre}
    \end{subfigure}	
    \begin{subfigure}{0.48\textwidth}
         \centering
         \includegraphics[width=\textwidth]{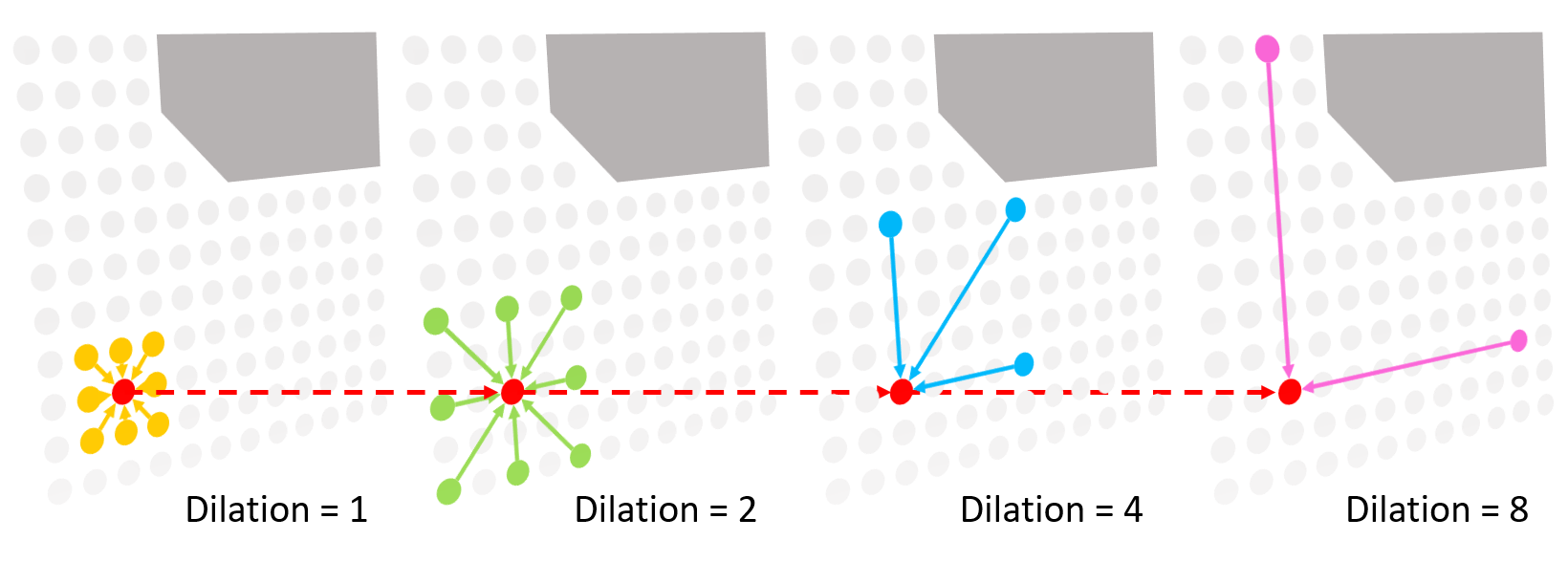}
         \caption{The graph with dilated connections for the DA layer. The gray area represents static obstacles in the driving environment, and the graph is constructed with constraints of the freespace. The target node is in red, while its dilated neighbors are in different colors.}
         \label{fig:da_dilated}
    \end{subfigure}
	\caption{Proposed graph neural network for the DA layer.}\label{fig:da_net}
	\vspace{-0.8cm}
\end{figure}

\subsubsection{DA graph encoder}
Recall that we formulate the DA layer as an 8-connected graph, i.e., a node is connected with its nearest neighbors in 8 directions (if any). The reason is that we expect that the node features can be updated by aggregating information only from neighboring nodes since the local connection restricts the operations to a small range, bringing higher computational efficiency. Inspired by the point cloud learning methods~\cite{qi2017pointnet}, we design a feature fusion method with permutation-invariant operations since the order of the neighboring nodes is not supposed to impact the aggregation results. A detailed illustration is presented in Fig.~\ref{fig:da_aggre}. For a target node, we extract the features of its neighbors using a multi-layer perceptron (MLP) whose weights are shared for all node features. Then, we use a max-pooling operation to get the global feature, which is concatenated with the target node's feature and further processed by another MLP. Residual connection~\cite{he2016deep} is also incorporated to get better performance.

A remaining problem for the aforementioned network is that the receptive field is limited, even if we stack multiple layers, since features are only exchanged between adjacent nodes. To solve this issue, we introduce multi-scale context fusion via dilated connections, which is widely used in computer vision~\cite{chen2014semantic, yu2015multi} and audio tasks~\cite{oord2016wavenet}. In Fig.~\ref{fig:da_dilated}, we demonstrate the dilated feature aggregation on the DA graph. The dilated connections are built in 8 directions with a dilation factor of $2^{l}, l\in[0,L)$, where $L=6$ is the number of layers. Specifically, for the $l$-th layer, we denote the input feature of the $i$-th node as $\mathbf{u}^l_i$, and the features of its neighboring nodes as $\mathbf{u}^l_j, j\in\mathcal{N}(i,l)$, where $\mathcal{N}(i,l)$ is a finite set of the neighbors' indices of the $i$-th node in layer $l$. We perform feature aggregation in a layer-by-layer manner, and the per-node update of the $l$-th layer can be written as
\begin{equation*}
    \mathbf{u}^{l+1}_i = \sigma(\mathbf{u}^{l}_i + \sigma((\mathbf{u}^{l}_i \boxplus \underset{j\in\mathcal{N}(i,l)}{\texttt{Max}}\{\sigma(\mathbf{u}^{l}_j W^{l}_1)\})W^{l}_2)W^{l}_3),
\end{equation*}
where $\{W^l_1, W^l_2, W^l_3\}$ are weights for fully-connected networks, $\sigma(\cdot)$ is a non-linear function consisting of LayerNorm~\cite{ba2016layer} and ReLU~\cite{nair2010rectified} activation, $\boxplus$ is the concatenation, and $\texttt{Max}\{\cdot\}$ is the global max-pooling operator. We consider this $L$-layered multi-scale fusion as a block, and denote it as $f_{\text{da}}:\mathbb{R}^{N_{\text{da}}\times d_{\text{da}}}\rightarrow \mathbb{R}^{N_{\text{da}}\times d_{\text{da}}}$, where $N_{\text{da}}$ is the number of nodes of the DA graph. Moreover, multiple blocks can be stacked to further enlarge the receptive field.

\subsubsection{LS graph encoder}\label{sec:ls_enc}

For the LS layer, it is important to consider long-range and topological features as traffic participants are strongly guided by this information. However, due to the special structure of lane graphs, it is non-trivial to apply general GNNs as the feature extraction backbone. For example, popular GNNs~\cite{kipf2016semi, velivckovic2017graph, hamilton2017inductive} mainly focus on the node feature without utilizing edge features adequately (e.g., types of connectivity). In this paper, we leverage the dilated \texttt{LaneConv} operator which was originally proposed in LaneGCN~\cite{liang2020learning}. The dilated \texttt{LaneConv} can effectively extract map features according to the lane graph topology and use dilated connections to capture the long-range dependency. We simply follows the original design of \texttt{LaneConv} and utilize it as the feature extractor of the LS layer. And the process can be represented as $f_{\text{ls}}:\mathbb{R}^{N_{\text{ls}}\times d_{\text{ls}}} \rightarrow \mathbb{R}^{N_{\text{ls}}\times d_{\text{ls}}}$, where $N_{\text{ls}}$ is the number of nodes in the LS layer. For detailed implementation, we refer interested readers to~\cite{liang2020learning}.

Although \texttt{LaneConv} is quite suitable to handle the lane graph topology, it is noticeable that the efficacy degenerates in some scenarios where lanes lack direct connection but are geometrically close (e.g., lane segments at intersections are topologically distant but geometrically overlapping). Fortunately, our DA layer effectively fills this gap since it provides alternative paths for the message passing among nodes. 

\subsection{GAT-based Message Passing}\label{sec:inter}
To achieve efficient message passing among the agents, DA and LS layers, a network based on graph attention is introduced. Features are aggregated to the target node from its neighbors (context) according to their importance. We denote the target node as $\mathbf{u}_i$ and its context node as $\mathbf{u}_j, j\in\mathcal{N}(i)$. In practice, context nodes are queried according to the Euclidean distance to the target node. We use $\alpha_{ij}$ as the attention weight indicating the importance of $\mathbf{u}_j$ to $\mathbf{u}_i$, which can be written as
\begin{equation*}
    \alpha_{ij} = \frac{\exp{(\phi((\mathbf{u}_iW_{\text{tgt}} \boxplus \mathbf{u}_jW_{\text{ctx}})W_{\text{att}}))}}{\sum_{k\in\mathcal{N}(i)}\exp{(\phi((\mathbf{u}_iW_{\text{tgt}} \boxplus \mathbf{u}_kW_{\text{ctx}})W_{\text{att}}))}},
\end{equation*}
where $\{W_{\text{tgt}},W_{\text{ctx}},W_{\text{att}}\}$ are trainable weights for linear projection, and $\phi$ is a non-linear function with normalization and LeakyReLU~\cite{maas2013rectifier}. Then, the aggregated result can be calculated using the weighted sum of the context nodes' features. The per-node update rule is written as
\begin{equation*}
    \mathbf{u}_i \leftarrow  \phi((\mathbf{u}_i \boxplus \sum_{j\in\mathcal{N}(i)}\alpha_{ij}\mathbf{u}_jW_1)W_2),
\end{equation*}
where $W_1\in\mathbb{R}^{d_{\text{ctx}}\times d_{\text{tgt}}}$ and $W_2\in\mathbb{R}^{(2\cdot d_{\text{tgt}})\times d_{\text{tgt}}}$ are the weights of fully-connected layers, while $d_{\text{tgt}}$ and $d_{\text{ctx}}$ are dimensions of the target node and context node features. Note that the residual connection is also incorporated and we omit it for brevity in the equation above.

\subsection{Multi-modal Goal Decoder}
After the dual-scale feature fusion, we perform the per-node classification on the DA graph, obtaining a heatmap that describes the probability distribution of the potential goal. Given the heatmap, some methods utilize a post-processing module to generate final predicted goals, such as non-maximum suppression (NMS)~\cite{gilles2021home} and optimization-based algorithms~\cite{gilles2021home, gu2021densetnt}. Such methods usually depend on human heuristics and bring more effort in parameter tuning. Here, we present a simple network to achieve accurate goal prediction. With a slight abuse of notation, given the generated heatmap, we select the top $S$ candidates with the highest scores and get their corresponding features, predicted scores, and coordinates. To balance the computation time and performance, we set the number of candidates $S=200$. We concatenate these vectors and use an MLP to get the new embedding for the selected nodes, and employ a self-attention mechanism~\cite{vaswani2017attention} to fuse the global heatmap information. Then, a multi-modal goal decoder is introduced, which contains $K=6$ headers with an identical network structure to extract different driving modes. For the $k$-th header, all node features are passed through a 1-D convolution-based network followed by a Softmax function to obtain the assignment scores $\gamma_{k,s}$, and the corresponding predicted goal $\hat{g}_k$ can be simply obtained by the weighted sum of the candidates' coordinate $p_{s}$, i.e., $\hat{g}_k = \sum^{S}_{s=1}\gamma_{k,s}p_{s}$. Intuitively, we expect each of the headers to generate a uni-modal prediction, and the mixture of all output can faithfully capture the multi-modality. The efficacy of our multi-modal goal decoder is evaluated in Sec.~\ref{sec:ablation}.

\subsection{Trajectory Completion}\label{sec:traj_compl}
Before the full trajectory decoding, the map context are aggregated to agent features, and interactions among agents are further fused using an GAT-based module (similar to the aggregator in Sec.~\ref{sec:inter}). Then, the completed trajectories are generated using an MLP with a residual block conditioned on predicted goals and agent features. For the confidence score of the predicted trajectory, we directly use the heatmap score of the DA node which is closest to the endpoint of the predicted trajectory. Despite the simple design, the proposed method is able to generate accurate and reasonable results.

\begin{table*}[tb]
	\centering
	\caption{Results on the Argoverse motion forecasting dataset (test split). Brier-minFDE is the official ranking metric.\label{tab:benchmark}}
	\setlength{\tabcolsep}{5mm}{
	\begin{tabular}{l|ccc|ccc >{\columncolor[gray]{0.9}}c}
	\toprule
	\multirow{2}{*}{Method}                 & \multicolumn{3}{c}{K=1} & \multicolumn{4}{c}{K=6}              \\
	                                        & minADE & minFDE & MR    & minADE & minFDE & MR  & Brier-minFDE \\
	\midrule
	Argo-NN baseline~\cite{chang2019argoverse}       & 3.46          & 7.88         & 87.2           & 1.71          & 3.29          & 53.7          & 3.98        \\
    LaneGCN~\cite{liang2020learning}	    & 1.71	        & 3.78         & 59.1           & 0.87          & 1.36          & 16.3          & 2.06        \\
    mmTransformer~\cite{liu2021multimodal}	& 1.77	        & 4.00         & 61.8           & 0.84          & 1.34          & 15.4          & 2.03        \\
    LaneRCNN~\cite{zeng2021lanercnn}	    & 1.68      	& 3.69         & \textbf{56.9}  & 0.90          & 1.45          & 12.3          & 2.15        \\
    PRIME~\cite{song2021learning}	        & 1.91	        & 3.82         & 58.7           & 1.22          & 1.56          & 11.5          & 2.10        \\
    DenseTNT~\cite{gu2021densetnt}	        & 1.68	        & \textbf{3.63}& 58.4           & 0.88          & 1.28          & 12.6          & 1.98        \\
    HOME~\cite{gilles2021home}              & 1.69          & 3.65         & 57.1           & 0.93          & 1.44          & 9.8           & 1.96        \\
    GOHOME~\cite{gilles2021gohome}	        & 1.69	        & 3.65         & 57.2           & 0.94          & 1.45          & 10.5          & 1.98        \\
    HO+GO~\cite{gilles2021gohome} (Ensemble)& 1.70          & 3.68         & 57.2           & 0.89          & 1.29          & \textbf{8.4}  & 1.86        \\
    \midrule
    DSP (ours)                              & \textbf{1.678}& 3.71         & 57.5           & \textbf{0.82} & \textbf{1.22} & 13.0          & \textbf{1.858}\\
  \bottomrule
\end{tabular}}
\vspace{-0.4cm}
\end{table*}

\subsection{Loss Function}\label{sec:learning}
Since we decompose the trajectory prediction into several sub-tasks, multiple losses are involved in the training phase.
\subsubsection{Goal classification}
The endpoint classification is essentially a multi-class classification problem with extremely unbalanced labels. To overcome this issue, we follow the method in~\cite{law2018cornernet}, which uses soft labels generated from a 2-D Gaussian kernel centered at the GT goal, and the objective is formulated as a modified focal loss~\cite{lin2017focal}:
\begin{equation*}
  \mathcal{L}_{\text{GC}} = -\frac{1}{N_p}\sum^{N_{\text{da}}}_{i}
    \begin{cases}
      (1-\hat{h}_i)^\alpha\log(\hat{h}_i)               & \text{if $h_i=1$}\\
      (1-h_i)^\beta (\hat{h}_i)^\alpha \log(1-\hat{h}_i)& \text{otherwise,}
    \end{cases}       
\end{equation*}
where $\alpha=2$, $\beta=4$ are parameters of the focal loss; $\hat{h}_i$ and $h_i$ are heatmap scores and GT labels; and $N_p$ is the number of positive labels. We define positive labels as the DA nodes whose distance from the GT goal is smaller than $1$ meter. 

\subsubsection{Goal regression}
For the multi-modal goal decoder, we perform a winner-takes-all (WTA) strategy to alleviate the mode collapse issue. Specifically, regression loss is only calculated for the predicted goal that has the minimum displacement to the GT endpoint:
\begin{equation*}
    \mathcal{L}_{\text{GR}} = \texttt{SmoothL1Loss}(g, \hat{g}_{k^*}),
\end{equation*}
where $k^*$ is the index of the best predicted goal.

\subsubsection{Trajectory regression}
Since there is only one GT trajectory for each sample, we apply a teacher forcing technique~\cite{williams1989learning} that uses the GT goal as the input of the trajectory completion network. The loss term is written as:
\begin{equation*}
    \mathcal{L}_{\text{TR}} = \texttt{SmoothL1Loss}(\mathbf{y}, \hat{\mathbf{y}}),
\end{equation*}
where $\mathbf{y}$ and $\hat{\mathbf{y}}$ are the GT future trajectory and predicted trajectory, respectively.

To summarize, the total loss function can be written as a linear combination of individual loss terms:
\begin{equation*}
    \mathcal{L} = \omega_1(\omega_2\mathcal{L}_{\text{GC}} + (1-\omega_2)\mathcal{L}_{\text{GR}}) + (1-\omega_1)\mathcal{L}_{\text{TR}}.
\end{equation*}
In practice, we set $\omega_1=\omega_2=0.8$ to address the importance of the endpoint classification. With the proposed objectives, our DSP can be trained in an end-to-end fashion.

\section{Experimental Results}\label{sec:experiments}

\subsection{Experiment Setup}
\subsubsection{Dataset}
We train and evaluate our model on the large-scale Argoverse motion forecasting dataset~\cite{chang2019argoverse}, which includes 205,942 sequences for training, 39,472 sequences for validation, and 78,143 sequences for testing. The dataset contains HD maps, and each sequence is sampled at 10 Hz, and the models are required to output trajectories for the target agent in the future 3 seconds according to the 2-second past observations (i.e., $T=20, H=30$).

\subsubsection{Metrics}
We follow the official evaluation metrics of the Argoverse benchmark and focus on several important ones that are most considered in recent works~\cite{liang2020learning, gilles2021home, zeng2021lanercnn, gu2021densetnt}. The average displacement error (ADE) is the average Euclidean distance between the predicted and GT trajectories, while the final displacement error (FDE) only calculates the error at endpoints. Considering multi-modal trajectory prediction, the minimum ADE (minADE) and minimum FDE (minFDE) are used to evaluate only the best forecast trajectory to the GT. Brier-minFDE and Brier-minADE add an additional Brier score $(1-p)^2$ to the L2 distances, where $p$ denotes the probability of the best forecast trajectory. The miss rate (MR) is the percentage of sequences where the obtained minFDE is greater than $2.0$ meters. We refer interested readers to~\cite{chang2019argoverse} for detailed definitions.

\subsubsection{Implementation details}
We train our model on 8 TITAN Xp GPUs using a batch size of 96 with the Adam~\cite{kingma2014adam} optimizer for 30 epochs. We set the learning rate as 1e-3 at the beginning and gradually decrease it to 1e-4 after 25 epochs. Data augmentation techniques, including random rotation and flipping, are applied during training. All the results are based on a single model, namely, no ensemble methods are used in either training or evaluation phases.

\subsection{Comparison with the State-of-the-art}
We compare our DSP with other state-of-the-art methods on the Argoverse motion forecasting benchmark.\footnote{https://eval.ai/web/challenges/challenge-page/454/leaderboard/1279} The quantitative results on the test set are shown in Table~\ref{tab:benchmark}. Note that we also report single model results for HOME~\cite{gilles2021home} and GOHOME~\cite{gilles2021gohome} for a fair comparison. The current official ranking metric is Brier-minFDE (K=6), which reflects both the multi-modal endpoint accuracy and probability estimation. Our DSP outperforms the listed state-of-the-art methods on many metrics, and even achieves similar accuracy to ensembled models (e.g., HO+GO~\cite{gilles2021gohome}), showing a strong ability in accurate motion forecasting. We point out that our method is also suitable for model ensembling since ensembling on a heatmap can significantly improve performance without any risk of mode collapse~\cite{gilles2021gohome}. We leave this as a future work.

\begin{figure}[t]
	\centering
	\begin{subfigure}{0.48\textwidth}
	    \centering
		\includegraphics[width=\textwidth]{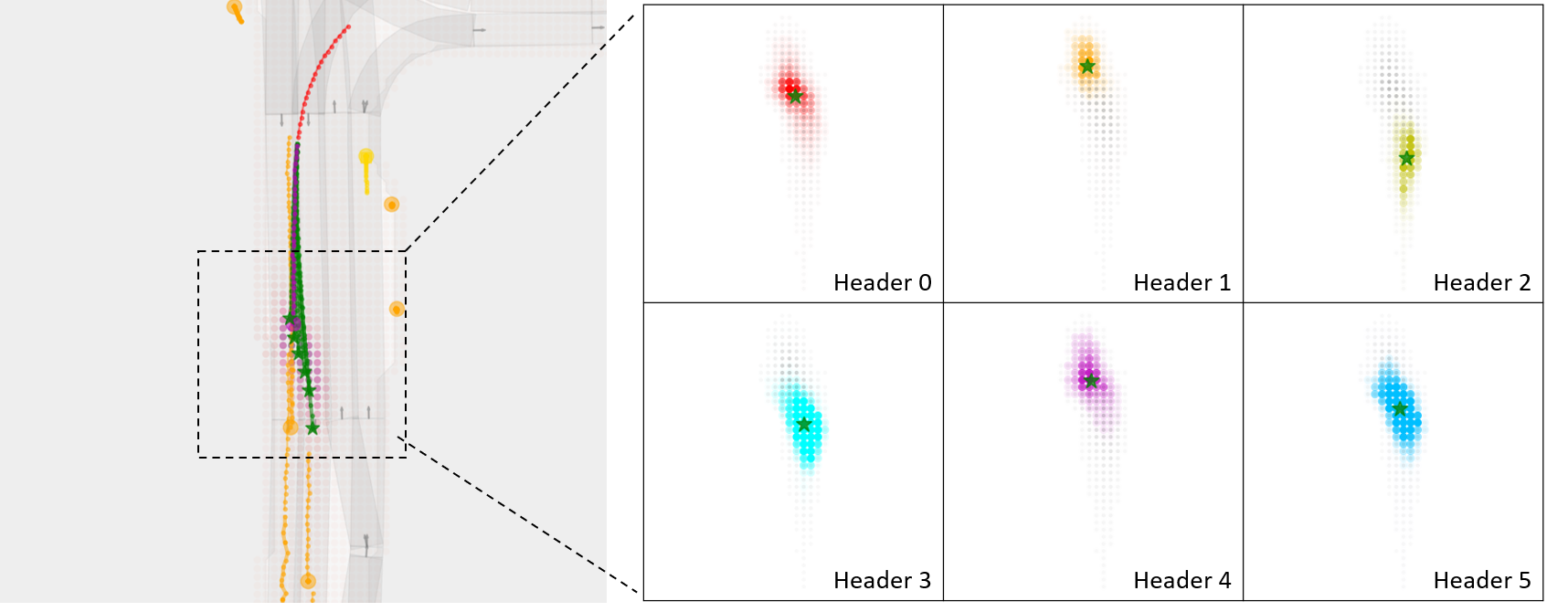}
	\end{subfigure}%
    \vspace{+0.1cm}
	\begin{subfigure}{0.48\textwidth}
	    \centering
		\includegraphics[width=\textwidth]{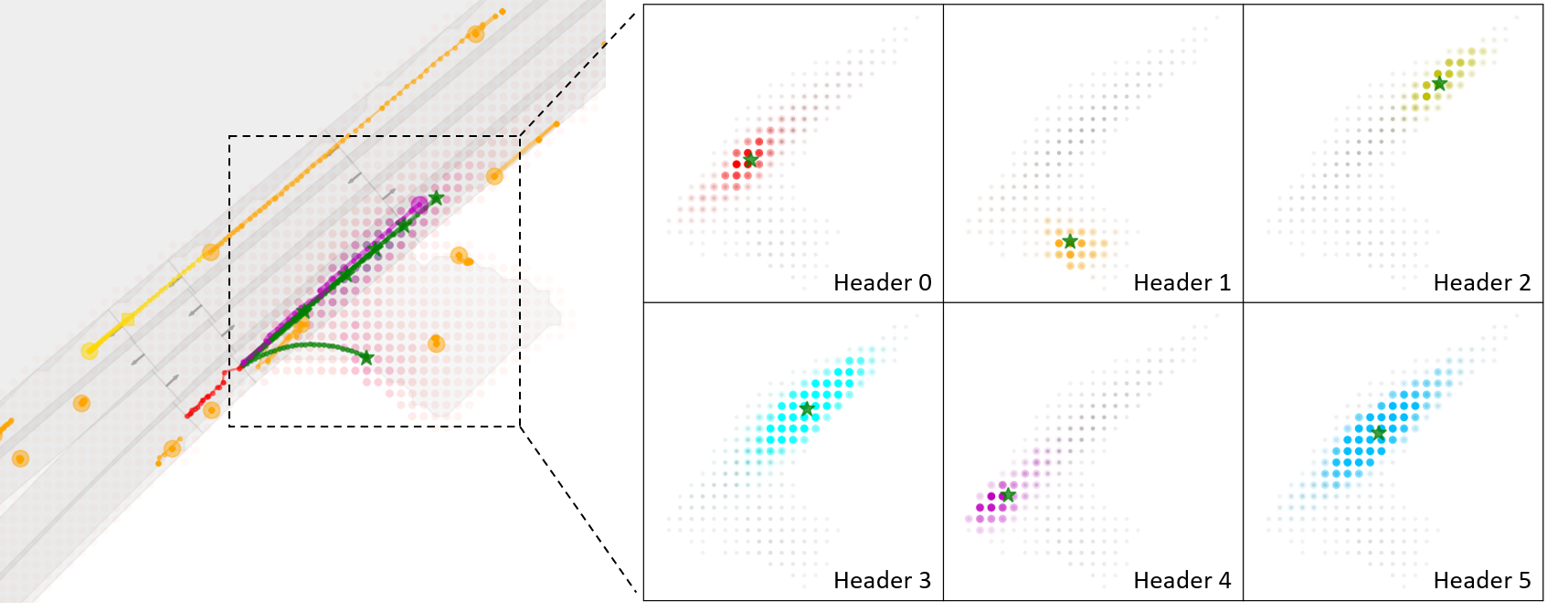}
	\end{subfigure}%
	\caption{Illustration of the multi-modal goal prediction (best viewed in color). The left-hand side shows the driving scenario. The observed history of the target agent is in red, while the ground truth future trajectory is shown in magenta. Multi-modal trajectories generated by the model are in green. Purple dots are sampled goal candidates, while the brightness represents the heatmap score. Trajectories of other agents are shown in orange. The right-hand side shows the assignment scores generated by headers of the decoder, and predicted goals are represented using green stars. Grey dots denote the high-scoring DA nodes of the predicted heatmap.}\label{fig:nn_dec}
	\vspace{-0.2cm}
\end{figure}

\begin{figure*}[t]
	\centering
	\begin{subfigure}{0.25\textwidth}
	    \centering
		\includegraphics[width=0.98\textwidth]{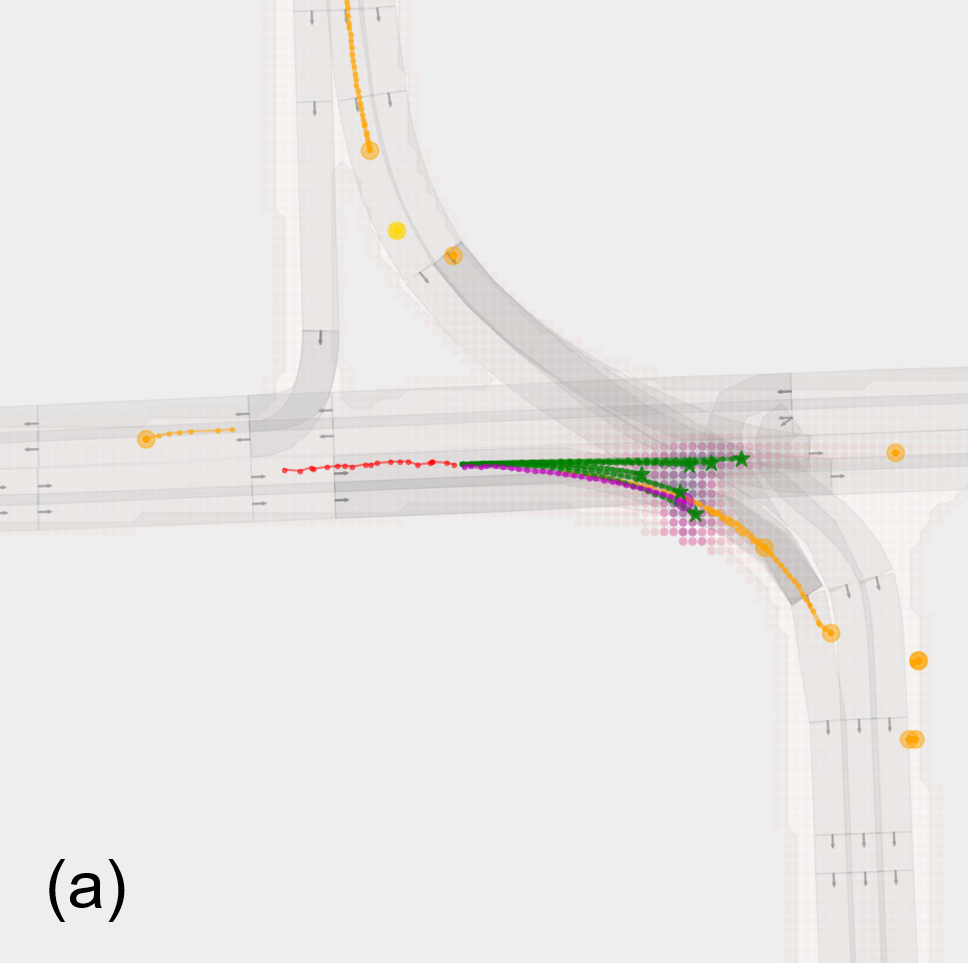}
	\end{subfigure}%
	\begin{subfigure}{0.25\textwidth}
	    \centering
		\includegraphics[width=0.98\textwidth]{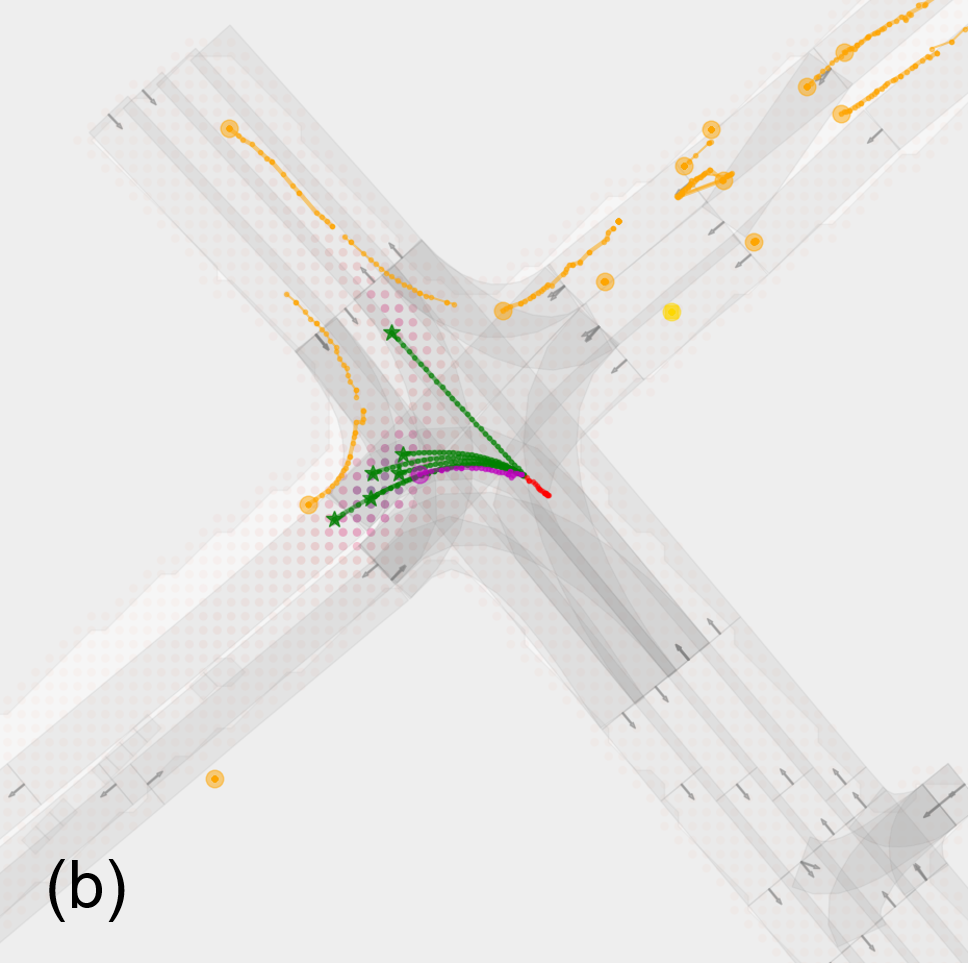}
	\end{subfigure}%
	\begin{subfigure}{0.25\textwidth}
	    \centering
		\includegraphics[width=0.98\textwidth]{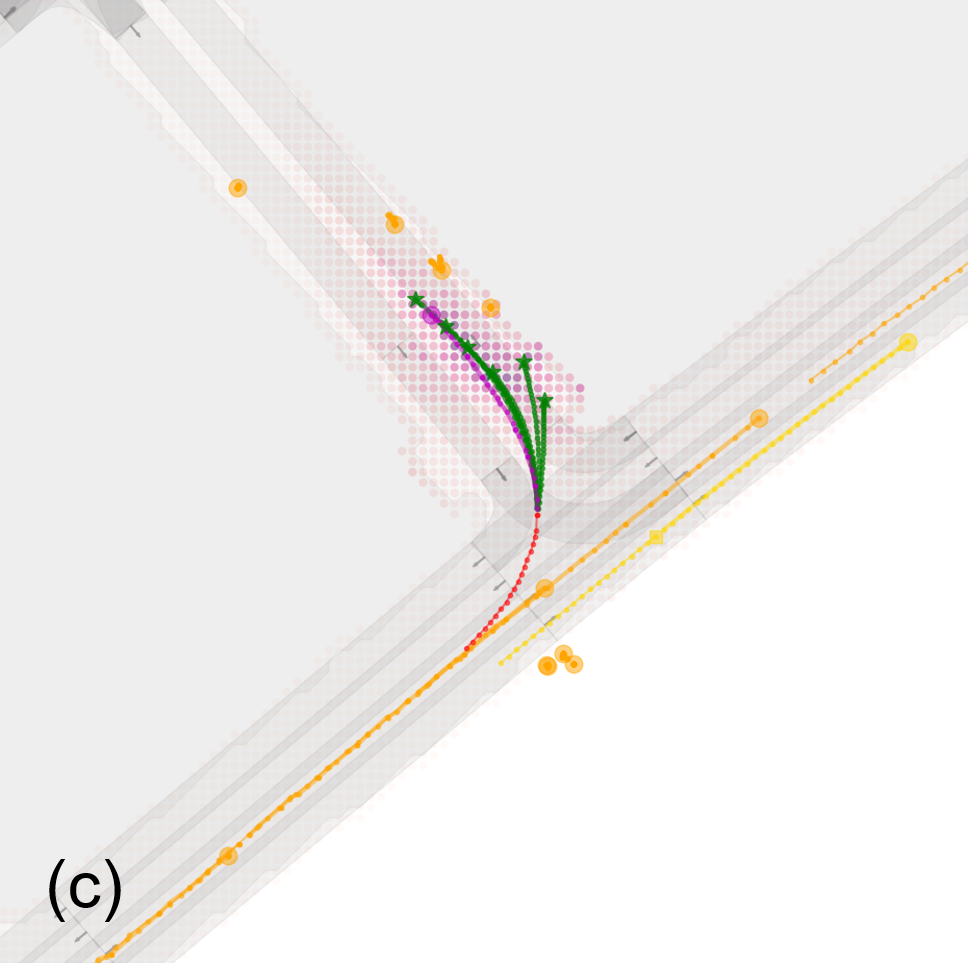}
	\end{subfigure}%
	\begin{subfigure}{0.25\textwidth}
	    \centering
		\includegraphics[width=0.98\textwidth]{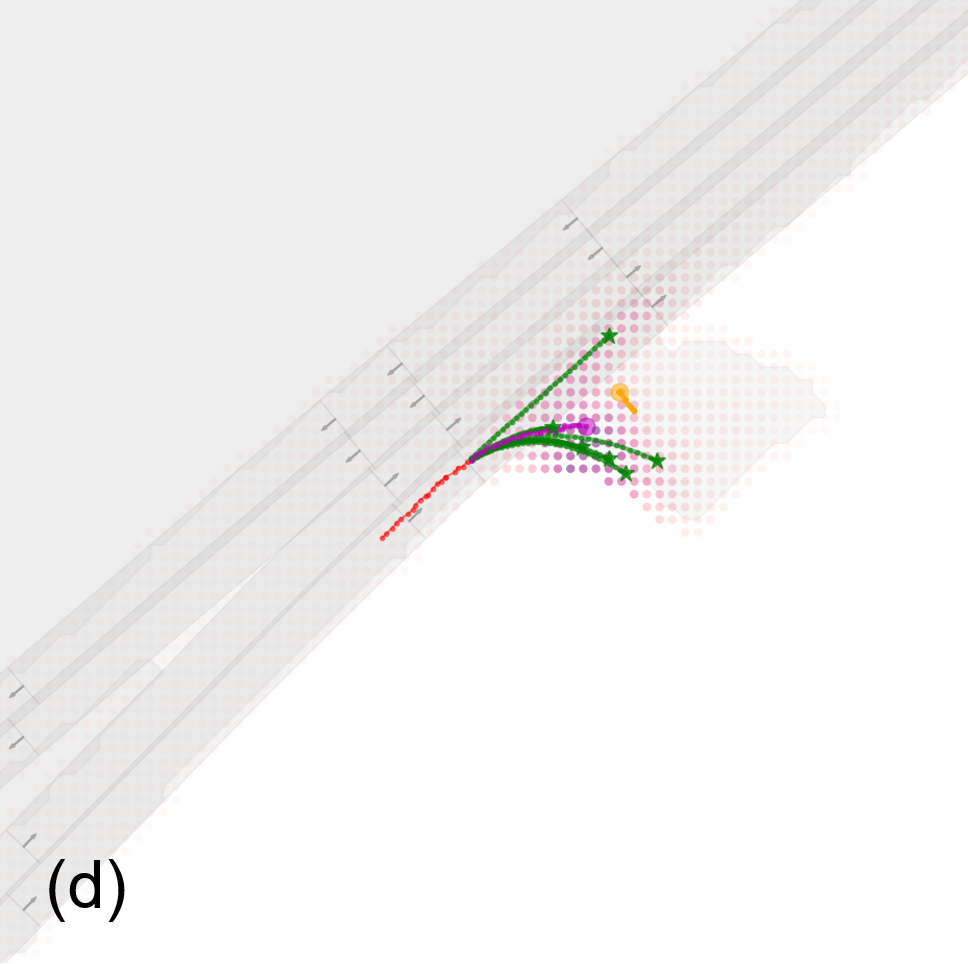}
	\end{subfigure}%
	\caption{Qualitative results of our DSP in various driving scenarios (The color scheme is identical to Fig.~\ref{fig:nn_dec}.). Scenario (a) and (b): our method accurately captures the multi-modal behavior when the target agent approaches an intersection. Scenario (c): the target agent is turning left, however, which is not allowed since no left-turn lane exists in the intersection. Our method captures the motion and makes correct prediction results. Scenario (d): the target agent is decelerating and driving off the road, which is hard to predict for methods that use lane information only. Our method can draw samples in the DA and capture the abnormal behaviors.}\label{fig:qualitative}
	\vspace{-0.4cm}
\end{figure*}

\subsection{Ablation Study}\label{sec:ablation}
\subsubsection{On the dual-scale context encoder}
To demonstrate the effectiveness of the proposed dual-scale feature fusion mechanism, we modify the network and obtain several variants. Evaluation results for these model variants on the Argoverse validation split are shown in Table~\ref{tab:ablation_enc}. We first show the benefit of the spatio-temporal occupancy feature of the DA node (S-T Occ.). Then, we investigate whether the fine-grained features are helpful for motion prediction by removing the feature aggregation process from the DA layer to the LS layer (DA2LS). Intuitively, the resulting variant is similar to DenseTNT~\cite{gu2021densetnt}, as both of the models fuse high-level features on a sparse graph and propagate features to dense endpoint samples to generate the goal heatmap. However, the differences are two-folds: 1) we use \texttt{LaneConv} for sparse feature aggregation while DenseTNT fuses features on a complete graph; 2) DenseTNT uses a global attention mechanism to pass messages from the sparse graph to dense nodes, while we use a distance-guided GAT, which is more efficient for computation and memory usage. For the proposed GAT-based inter-layer aggregator, we implement a module based on the global max-pooling as an alternative. From Table~\ref{tab:ablation_enc} we can observe that all proposed modules improve the prediction performance, showing the effectiveness of the proposed context encoder.

\begin{table}[t]
	\centering
	\caption{Ablation study on the dual-scale context encoder on the Argoverse motion forecasting dataset (validation split).\label{tab:ablation_enc}}
	\setlength{\tabcolsep}{1.2mm}{
	\begin{tabular}{@{}ccc|cccc@{}}
	\toprule
	S-T Occ.   & DA2LS & \makecell*[c]{Inter-layer\\ Aggre.} & minADE & minFDE & MR  & Brier-minFDE \\
	\midrule
	           &            & GAT       & 0.72 & 1.03 & 9.6 & 1.64     \\
	           & \checkmark & GAT       & 0.71 & 1.02 & 9.8 & 1.64     \\
	\checkmark &            & GAT       & 0.71 & 1.02 & 9.6 & 1.63     \\
	\checkmark & \checkmark & MaxPool   & 0.72 & 1.02 & 9.4 & 1.64     \\
	\checkmark & \checkmark & GAT       & \textbf{0.69} & \textbf{0.98} & \textbf{9.0} & \textbf{1.61}     \\
  \bottomrule
\end{tabular}}
\vspace{-0.8cm}
\end{table}

\subsubsection{On the goal decoder}
Once the endpoint classification score is obtained, various methods can be applied to generate the final goal hypothesis. For comparison, an NMS-based goal selection strategy is implemented. We sort the goal candidates according to their scores in descending order, and then select the goals into a set from the top of the queue in a greedy manner. The newly picked candidates are required to be at least $R_{\text{supp}}$ away from all points in the set, and are otherwise dropped. If a sufficient number of candidates is not obtained, we decay $R_{\text{supp}}$ by multiplying it by a discount factor $\kappa$ and validate the candidates from the top of the queue again until enough goal candidates are collected. In practice, we set $R_{\text{supp}}=2.8$ meters and $\kappa=0.8$ for obtaining better performance. We also employ weighted k-means clustering-based goal generation as another baseline, and the input of the k-means algorithm is a set of 2-D coordinates with heatmap scores as corresponding weights. 
The evaluation results are shown in Table~\ref{tab:ablation_post}. We can find that the NMS-based goal selection yields better coverage of the high-probability region due to its greedy property resulting in a lower MR, while our NN-based goal decoder directly optimizes minFDE, leading to lower displacement error. Previous works such as \cite{gilles2021home} and \cite{gu2021densetnt} propose post-processing methods to further decrease the displacement error. However, these methods are always separated from the prediction network and are not differentiable. In contrast, our NN-based goal decoder is able to back-propagate the goal prediction error thanks to the end-to-end training, which significantly improves the prediction accuracy and even outperforms the well-designed post-processing methods based on the online and offline optimization (see Table~\ref{tab:ablation_post}).

\begin{table}[t]
	\centering
	\caption{Ablation study on the different types of goal decoders on the Argoverse motion forecasting dataset (validation set).\label{tab:ablation_post}}
	\begin{tabular}{l|cccc}
	\toprule
	Method & minADE & minFDE & MR \\
	\midrule
	HOME (FDE L=6)~\cite{gilles2021home}                & -             & 1.16          & \textbf{7.4}  \\
	DenseTNT 100ms opt. (minFDE)~\cite{gu2021densetnt}  & 0.73          & 1.05          & 9.8           \\
	\midrule
    DSP + NMS                                           & 0.76          & 1.22          & 7.6           \\
	DSP + K-means                                       & 0.70          & 1.04          & 9.0           \\
	DSP + NN                                            & \textbf{0.69} & \textbf{0.98} & 9.0           \\
  \bottomrule
\end{tabular}
\vspace{-0.8cm}
\end{table}

In Fig.~\ref{fig:nn_dec}, we provide an example to show the multi-modal output of our NN-based goal decoder. Intuitively, the heatmap represents the probability distribution of the target's goal, which is highly multi-modal. Our decoder is expected to ``decompose" the multi-modal distribution into several uni-modal distributions. On the right-hand side of Fig.~\ref{fig:nn_dec}, we illustrate the assignment scores (represented using the transparency) generated by different headers. We can find that the multi-modality is well captured by the decoder, as different headers focus on different parts of the heatmap. Another interesting observation is that one header tends to generate similar behavior in different scenarios. For example, in Fig.~\ref{fig:nn_dec}, Header 2 prefers aggressive behavior (higher speed), while Header 4 is more conservative (deceleration) in both of the given scenarios. This shows that our DSP is able to learn different driving modalities in an implicit manner.

\begin{table}[t]
	\centering
	\caption{Inference time for a single target agent w.r.t. the maximum map size on the Argoverse dataset (validation split).\label{tab:ablation_time}}
	\setlength{\tabcolsep}{1.2mm}{
	\begin{tabular}{l|cccc}
	\toprule
	Max. map size ($m$)    & 120    & 160    & 200     & 240      \\
	\midrule
	\# nodes of DA layer    & 3.2K   & 4.6K   & 6.1K   & 7.8K     \\
	\# nodes of LS layer    & 0.50K  & 0.63K  & 0.77K  & 0.93K    \\
	\midrule
	Encoding time ($ms$)  & 7.2    & 9.0    & 11.3    & 13.3     \\
	Decoding time ($ms$)  & 0.83   & 0.82   & 0.83    & 0.85     \\
	Total time ($ms$) / FPS (Hz)   & 9.3 / 108    & 11.3 / 88  & 13.5 / 74     & 15.7 / 64     \\
  \bottomrule
\end{tabular}}
\vspace{-0.8cm}
\end{table}

\subsubsection{On the time-consuming}
Computational time is also an essential factor for the real-world automated driving system. To study the efficiency of the proposed method, we evaluate the network using maps with different maximum size. Note that the larger map size brings more graph nodes, leading to longer computational time. We conduct the experiment using a consumer PC with a single GPU, and the result is shown in Table~\ref{tab:ablation_time}. We can find that the inference speed is far beyond the real-time requirement even without any acceleration technique. Moreover, the total inference time grows linearly with the number of nodes, which shows the high computational efficiency of the proposed method.

\subsection{Qualitative Results}
We present qualitative results on the Argoverse dataset in Fig.~\ref{fig:qualitative}. Our DSP is able to generate accurate and realistic future motion in various scenarios. As shown in Fig.~\ref{fig:qualitative}, (a) and (b) present multi-modal trajectory prediction conforming to the lane geometry in complex intersections, while (c) and (d) show reasonable prediction results even when the target agent does not follow the traffic rules. More qualitative results can be found in the attached video.

\section{Conclusion}\label{sec:conclusion}
In this paper, we proposed DSP, a hierarchical graph-based network for accurate multi-modal trajectory prediction. We propose a double-layered driving context representation, and perform feature extraction as well as aggregation using GNNs, leading to higher flexibility and data efficiency. The proposed multi-modal goal-decoding network further improves the prediction accuracy and outperforms other rule-based baselines. We also demonstrate that our DSP achieves state-of-the-art performance on the large-scale Argoverse motion forecasting benchmark. Currently, our method focuses on the prediction for a single target, while the joint prediction for multiple targets~\cite{ettinger2021large} is more practical for the downstream planning tasks. We will try to extend our method for scene-consistent multi-agent prediction in the future.


\begin{thebibliography}{10}
	\bibitem{lefevre2014survey}
	S.~Lef{\`e}vre, D.~Vasquez, and C.~Laugier, ``A survey on motion prediction and
	  risk assessment for intelligent vehicles,'' \emph{ROBOMECH journal}, vol.~1,
	  no.~1, pp. 1--14, 2014.
	
	\bibitem{mozaffari2020deep}
	S.~Mozaffari, O.~Y. Al-Jarrah, M.~Dianati, P.~Jennings, and A.~Mouzakitis,
	  ``Deep learning-based vehicle behavior prediction for autonomous driving
	  applications: A review,'' \emph{IEEE Transactions on Intelligent
	  Transportation Systems}, 2020.
	
	\bibitem{gao2020vectornet}
	J.~Gao, C.~Sun, H.~Zhao, Y.~Shen, D.~Anguelov, C.~Li, and C.~Schmid,
	  ``{VectorNet}: Encoding {HD} maps and agent dynamics from vectorized
	  representation,'' in \emph{Proc. of the CVPR}, 2020, pp. 11\,525--11\,533.
	
	\bibitem{liang2020learning}
	M.~Liang, B.~Yang, R.~Hu, Y.~Chen, R.~Liao, S.~Feng, and R.~Urtasun, ``Learning
	  lane graph representations for motion forecasting,'' in \emph{Proc. of the
	  ECCV}.\hskip 1em plus 0.5em minus 0.4em\relax Springer, 2020, pp. 541--556.
	
	\bibitem{mangalam2020not}
	K.~Mangalam, H.~Girase, S.~Agarwal, K.-H. Lee, E.~Adeli, J.~Malik, and
	  A.~Gaidon, ``It is not the journey but the destination: Endpoint conditioned
	  trajectory prediction,'' in \emph{Proc. of the ECCV}.\hskip 1em plus 0.5em
	  minus 0.4em\relax Springer, 2020, pp. 759--776.
	
	\bibitem{zhao2020tnt}
	H.~Zhao, J.~Gao, T.~Lan, C.~Sun, B.~Sapp, B.~Varadarajan, Y.~Shen, Y.~Shen,
	  Y.~Chai, C.~Schmid, \emph{et~al.}, ``{TNT}: Target-driven trajectory
	  prediction,'' \emph{arXiv preprint arXiv:2008.08294}, 2020.
	
	\bibitem{gilles2021home}
	T.~Gilles, S.~Sabatini, D.~Tsishkou, B.~Stanciulescu, and F.~Moutarde, ``Home:
	  Heatmap output for future motion estimation,'' \emph{arXiv preprint
	  arXiv:2105.10968}, 2021.
	
	\bibitem{zeng2021lanercnn}
	W.~Zeng, M.~Liang, R.~Liao, and R.~Urtasun, ``{LaneRCNN}: Distributed
	  representations for graph-centric motion forecasting,'' \emph{arXiv preprint
	  arXiv:2101.06653}, 2021.
	
	\bibitem{gilles2021gohome}
	T.~Gilles, S.~Sabatini, D.~Tsishkou, B.~Stanciulescu, and F.~Moutarde,
	  ``{GOHOME}: Graph-oriented heatmap output forfuture motion estimation,''
	  \emph{arXiv preprint arXiv:2109.01827}, 2021.
	
	\bibitem{cui2019multimodal}
	H.~Cui, V.~Radosavljevic, F.-C. Chou, T.-H. Lin, T.~Nguyen, T.-K. Huang,
	  J.~Schneider, and N.~Djuric, ``Multimodal trajectory predictions for
	  autonomous driving using deep convolutional networks,'' in \emph{Proc. of the
	  ICRA}.\hskip 1em plus 0.5em minus 0.4em\relax IEEE, 2019, pp. 2090--2096.
	
	\bibitem{alahi2016social}
	A.~Alahi, K.~Goel, V.~Ramanathan, A.~Robicquet, L.~Fei-Fei, and S.~Savarese,
	  ``Social {LSTM}: Human trajectory prediction in crowded spaces,'' in
	  \emph{Proc. of the CVPR}, 2016, pp. 961--971.
	
	\bibitem{deo2018convolutional}
	N.~Deo and M.~M. Trivedi, ``Convolutional social pooling for vehicle trajectory
	  prediction,'' in \emph{Proc. of the CVPR Workshops}, 2018, pp. 1468--1476.
	
	\bibitem{vemula2018social}
	A.~Vemula, K.~Muelling, and J.~Oh, ``Social attention: Modeling attention in
	  human crowds,'' in \emph{Proc. of the ICRA}.\hskip 1em plus 0.5em minus
	  0.4em\relax IEEE, 2018, pp. 4601--4607.
	
	\bibitem{chai2019multipath}
	Y.~Chai, B.~Sapp, M.~Bansal, and D.~Anguelov, ``{MultiPath}: Multiple
	  probabilistic anchor trajectory hypotheses for behavior prediction,''
	  \emph{arXiv preprint arXiv:1910.05449}, 2019.
	
	\bibitem{zhao2019multi}
	T.~Zhao, Y.~Xu, M.~Monfort, W.~Choi, C.~Baker, Y.~Zhao, Y.~Wang, and Y.~N. Wu,
	  ``Multi-agent tensor fusion for contextual trajectory prediction,'' in
	  \emph{Proc. of the CVPR}, 2019, pp. 12\,126--12\,134.
	
	\bibitem{phan2020covernet}
	T.~Phan-Minh, E.~C. Grigore, F.~A. Boulton, O.~Beijbom, and E.~M. Wolff,
	  ``{CoverNet}: Multimodal behavior prediction using trajectory sets,'' in
	  \emph{Proc. of the CVPR}, 2020, pp. 14\,074--14\,083.
	
	\bibitem{zhang2020map}
	L.~Zhang, P.-H. Su, J.~Hoang, G.~C. Haynes, and M.~Marchetti-Bowick,
	  ``Map-adaptive goal-based trajectory prediction,'' \emph{arXiv preprint
	  arXiv:2009.04450}, 2020.
	
	\bibitem{luo2020probabilistic}
	C.~Luo, L.~Sun, D.~Dabiri, and A.~Yuille, ``Probabilistic multi-modal
	  trajectory prediction with lane attention for autonomous vehicles,'' in
	  \emph{Proc. of the IROS}.\hskip 1em plus 0.5em minus 0.4em\relax IEEE, 2020,
	  pp. 2370--2376.
	
	\bibitem{ding2021epsilon}
	W.~Ding, L.~Zhang, J.~Chen, and S.~Shen, ``Epsilon: An efficient planning
	  system for automated vehicles in highly interactive environments,''
	  \emph{IEEE Transactions on Robotics}, 2021.
	
	\bibitem{song2021learning}
	H.~Song, D.~Luan, W.~Ding, M.~Y. Wang, and Q.~Chen, ``Learning to predict
	  vehicle trajectories with model-based planning,'' \emph{arXiv preprint
	  arXiv:2103.04027}, 2021.
	
	\bibitem{gu2021densetnt}
	J.~Gu, C.~Sun, and H.~Zhao, ``{DenseTNT}: End-to-end trajectory prediction from
	  dense goal sets,'' \emph{arXiv preprint arXiv:2108.09640}, 2021.
	
	\bibitem{velivckovic2017graph}
	P.~Veli{\v{c}}kovi{\'c}, G.~Cucurull, A.~Casanova, A.~Romero, P.~Lio, and
	  Y.~Bengio, ``Graph attention networks,'' \emph{arXiv preprint
	  arXiv:1710.10903}, 2017.
	
	\bibitem{qi2017pointnet}
	C.~R. Qi, H.~Su, K.~Mo, and L.~J. Guibas, ``{PointNet}: Deep learning on point
	  sets for 3d classification and segmentation,'' in \emph{Proc. of the CVPR},
	  2017, pp. 652--660.
	
	\bibitem{he2016deep}
	K.~He, X.~Zhang, S.~Ren, and J.~Sun, ``Deep residual learning for image
	  recognition,'' in \emph{Proc. of the CVPR}, 2016, pp. 770--778.
	
	\bibitem{chen2014semantic}
	L.-C. Chen, G.~Papandreou, I.~Kokkinos, K.~Murphy, and A.~L. Yuille, ``Semantic
	  image segmentation with deep convolutional nets and fully connected {CRFs},''
	  \emph{arXiv preprint arXiv:1412.7062}, 2014.
	
	\bibitem{yu2015multi}
	F.~Yu and V.~Koltun, ``Multi-scale context aggregation by dilated
	  convolutions,'' \emph{arXiv preprint arXiv:1511.07122}, 2015.
	
	\bibitem{oord2016wavenet}
	A.~V.~D. Oord, S.~Dieleman, H.~Zen, K.~Simonyan, O.~Vinyals, A.~Graves,
	  N.~Kalchbrenner, A.~Senior, and K.~Kavukcuoglu, ``{WaveNet}: A generative
	  model for raw audio,'' \emph{arXiv preprint arXiv:1609.03499}, 2016.
	
	\bibitem{ba2016layer}
	J.~L. Ba, J.~R. Kiros, and G.~E. Hinton, ``Layer normalization,'' \emph{arXiv
	  preprint arXiv:1607.06450}, 2016.
	
	\bibitem{nair2010rectified}
	V.~Nair and G.~E. Hinton, ``Rectified linear units improve restricted
	  {Boltzmann} machines,'' in \emph{Proc. of the ICML}, 2010.
	
	\bibitem{kipf2016semi}
	T.~N. Kipf and M.~Welling, ``Semi-supervised classification with graph
	  convolutional networks,'' \emph{arXiv preprint arXiv:1609.02907}, 2016.
	
	\bibitem{hamilton2017inductive}
	W.~L. Hamilton, R.~Ying, and J.~Leskovec, ``Inductive representation learning
	  on large graphs,'' in \emph{Proc. of the NeurIPS}, 2017, pp. 1025--1035.
	
	\bibitem{maas2013rectifier}
	A.~L. Maas, A.~Y. Hannun, A.~Y. Ng, \emph{et~al.}, ``Rectifier nonlinearities
	  improve neural network acoustic models,'' in \emph{Proc. of the ICML},
	  vol.~30, no.~1.\hskip 1em plus 0.5em minus 0.4em\relax Citeseer, 2013, p.~3.
	
	\bibitem{vaswani2017attention}
	A.~Vaswani, N.~Shazeer, N.~Parmar, J.~Uszkoreit, L.~Jones, A.~N. Gomez,
	  {\L}.~Kaiser, and I.~Polosukhin, ``Attention is all you need,'' in
	  \emph{Proc. of the NeurIPS}, 2017, pp. 5998--6008.
	
	\bibitem{chang2019argoverse}
	M.-F. Chang, J.~Lambert, P.~Sangkloy, J.~Singh, S.~Bak, A.~Hartnett, D.~Wang,
	  P.~Carr, S.~Lucey, D.~Ramanan, \emph{et~al.}, ``Argoverse: {3D} tracking and
	  forecasting with rich maps,'' in \emph{Proc. of the CVPR}, 2019, pp.
	  8748--8757.
	
	\bibitem{liu2021multimodal}
	Y.~Liu, J.~Zhang, L.~Fang, Q.~Jiang, and B.~Zhou, ``Multimodal motion
	  prediction with stacked transformers,'' in \emph{Proc. of the CVPR}, 2021,
	  pp. 7577--7586.
	
	\bibitem{law2018cornernet}
	H.~Law and J.~Deng, ``{CornerNet}: Detecting objects as paired keypoints,'' in
	  \emph{Proc. of the ECCV}, 2018, pp. 734--750.
	
	\bibitem{lin2017focal}
	T.-Y. Lin, P.~Goyal, R.~Girshick, K.~He, and P.~Doll{\'a}r, ``Focal loss for
	  dense object detection,'' in \emph{Proc. of the ICCV}, 2017, pp. 2980--2988.
	
	\bibitem{williams1989learning}
	R.~J. Williams and D.~Zipser, ``A learning algorithm for continually running
	  fully recurrent neural networks,'' \emph{Neural Computation}, vol.~1, no.~2,
	  pp. 270--280, 1989.
	
	\bibitem{kingma2014adam}
	D.~P. Kingma and J.~Ba, ``Adam: A method for stochastic optimization,''
	  \emph{arXiv preprint arXiv:1412.6980}, 2014.
	
	\bibitem{ettinger2021large}
	S.~Ettinger, S.~Cheng, B.~Caine, C.~Liu, H.~Zhao, S.~Pradhan, Y.~Chai, B.~Sapp,
	  C.~R. Qi, Y.~Zhou, \emph{et~al.}, ``Large scale interactive motion
	  forecasting for autonomous driving: The waymo open motion dataset,'' in
	  \emph{Proc. of the ICCV}, 2021, pp. 9710--9719.
	
\end{thebibliography}
\end{document}